\documentclass[10pt,twocolumn,letterpaper]{article}

\usepackage{cvpr}
\usepackage{times}
\usepackage{epsfig}
\usepackage{graphicx}
\usepackage{amsmath}
\usepackage{amssymb}
\usepackage{makecell}
\usepackage{multirow}
\usepackage{color}
\definecolor{green}{rgb}{0, 0.5, 0}
\definecolor{orange}{rgb}{0.8, 0.6, 0.2}
\definecolor{red}{rgb}{1.0, 0.0, 0.0}
\definecolor{teal}{rgb}{0.0, 0.4, 0.4}
\definecolor{purple}{rgb}{0.65,0,0.65}
\definecolor{saffron}{rgb}{0.95,0.75,0.2}
\definecolor{turquoise}{rgb}{0.0,0.5,0.5}
\definecolor{black}{rgb}{0.0, 0.0, 0.0}

\newcommand{\kx}[1]{{\color{black}#1}}

\usepackage[normalem]{ulem}

\usepackage{amsopn}
\usepackage{mathtools}
\usepackage{amsmath, amsthm, amsfonts, amssymb}
\usepackage{graphicx} 
\usepackage{textcomp}
\usepackage{xfrac}

\usepackage{overpic}
\usepackage{graphicx}
\usepackage{float}
\usepackage{multirow}

\usepackage[pagebackref=true,breaklinks=true,letterpaper=true,colorlinks,bookmarks=false]{hyperref}

\cvprfinalcopy 

\def\cvprPaperID{4313} 

\ifcvprfinal\fi
\begin{document}

\title{PartNet: A Recursive Part Decomposition Network for Fine-grained and Hierarchical Shape Segmentation}

\author{
Fenggen Yu$^{1\ast}$ \quad\quad Kun Liu$^{1}$\thanks{Joint first authors}
\quad\quad Yan Zhang$^1$ \quad\quad Chenyang Zhu$^2$ \quad\quad Kai Xu$^2$\thanks{Corresponding author: kevin.kai.xu@gmail.com}\\
$^1$Nanjing University \quad\quad $^2$National University of Defense Technology\\
}

\maketitle


\begin{abstract}
Deep learning approaches to 3D shape segmentation are typically formulated as a multi-class labeling problem.
Existing models are trained for a fixed set of labels, which greatly limits their flexibility and adaptivity.
We opt for top-down recursive decomposition and develop the first deep learning model for hierarchical segmentation of 3D shapes, based on recursive neural networks.
Starting from a full shape represented as a point cloud, our model performs recursive binary decomposition,
where the decomposition network at all nodes in the hierarchy share weights.
At each node, a node classifier is trained to determine
the type (adjacency or symmetry) and stopping criteria of its decomposition.
The features extracted in higher level nodes are recursively propagated to lower level ones.
Thus, the meaningful decompositions in higher levels provide strong contextual cues constraining the segmentations in lower levels.
Meanwhile, to increase the segmentation accuracy at each node, we enhance the recursive contextual feature
with the shape feature extracted for the corresponding part.
Our method segments a 3D shape in point cloud into an unfixed number of parts, depending on the shape complexity, showing strong generality and flexibility.
It achieves the state-of-the-art performance, both for fine-grained and semantic segmentation, on the public benchmark and a new benchmark of fine-grained segmentation proposed in this work.
We also demonstrate its application for fine-grained part refinements in image-to-shape reconstruction.
\end{abstract}


\section{Introduction}
Segmentation is a long-standing problem in 3D shape analysis, on which data-driven
approach has shown clear advantage over traditional geometric methods~\cite{xu2016data}.
With the proliferation of deep learning techniques, researchers have been seeking for
exploiting the powerful feature learning ability of deep neural networks to replace the hand-crafted
features used in previous data-driven approaches. In these works,
deep networks are trained for multi-class labeling task, which outputs a semantic label for each
geometric primitive (such as voxels~\cite{riegler2017octnet} or points~\cite{qi2017pnpp}).

\begin{figure}[t]
  \centering
  \includegraphics[width=\linewidth]{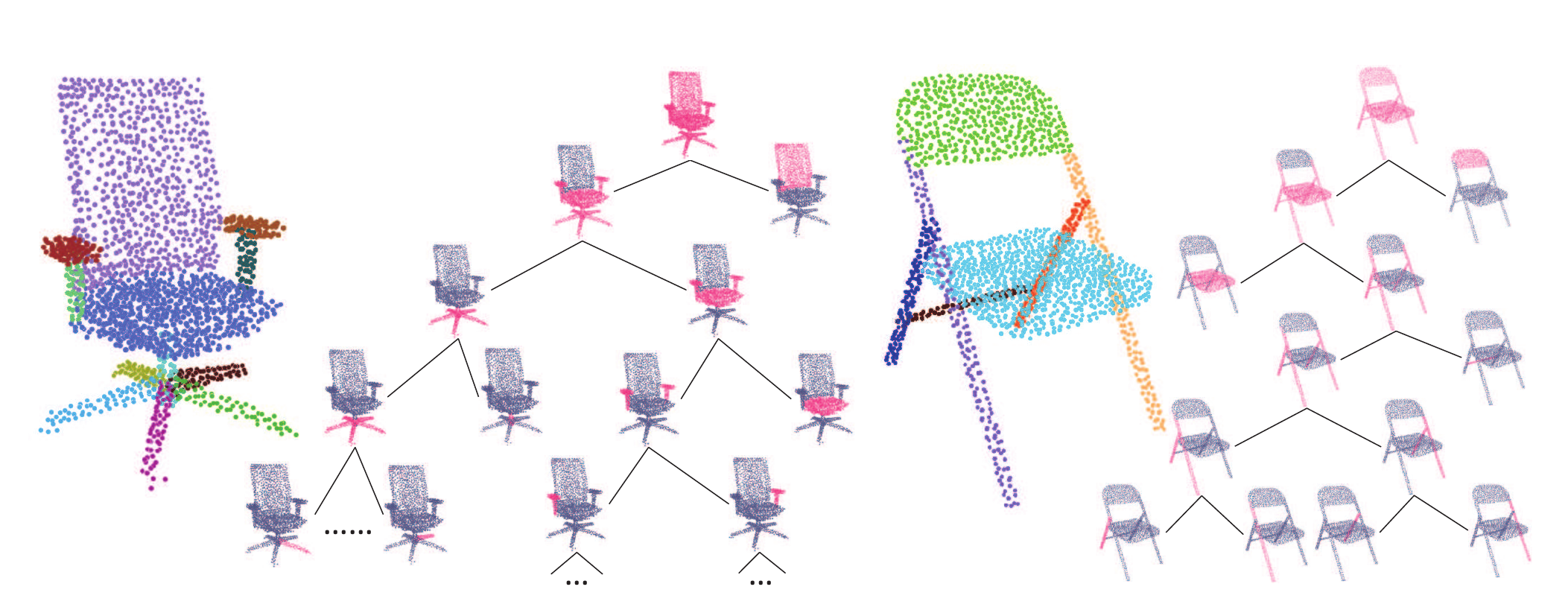}
  \caption{PartNet segments 3D point clouds in a top-down recursive fashion, leading to a hierarchy of fine-grained parts. The same model trained for Chair class can be used to segment different chair models into different number of parts, depending on the structure complexity of the input shapes.}
  \label{fig:teaser}
\end{figure}

There are two issues with these existing models.
First, the models are trained targeting a \emph{fixed} set of labels, which greatly limits
its flexibility and adaptivity. For example, a model trained to segment a chair into three semantic parts
cannot be used to correctly segment a chair with four parts, even they both belong to the same shape family.
Training different models for different targeted label sets is neither general nor efficient.
Second, labeling all primitives simultaneously cannot exploit the hierarchical nature of shape decomposition.
Hierarchical shape segmentation reduces the difficulty of shape segmentation through dividing the multi-class labeling problem into a cascade of binary labeling problems~\cite{attene2006hierarchical,Yi_SG17}.
On the other hand, hierarchical segmentation can utilize structural constraints across different levels:
The segmentations in higher levels provide strong cues constraining those in the lower levels.
This enables accurate segmentation into very fine-grained levels (Figure~\ref{fig:teaser}).

In this work, we opt for the top-down decomposition and propose the first deep learning model
for hierarchical segmentation of 3D shapes into fine-grained parts, based on recursive neural networks (RvNN).
Starting from a full shape represented as a point cloud, our model performs recursive binary decomposition,
where the decomposition network at all nodes in the hierarchy share weights.
At each node, a node classifier is trained to determine
the type of its decomposition (adjacency or symmetry node) and whether the decomposition should stop (leaf node).
The features extracted in higher level nodes are recursively propagated to lower level ones through the tree
structure, which we refer to as \emph{recursive context features}.
Therefore, the meaningful decompositions in higher levels constrain the segmentations in lower levels.
Meanwhile, to increase the segmentation accuracy at each node, we enhance the recursive context feature
with the \emph{part shape feature} extracted for the corresponding point cloud.

The network is trained with point sampled 3D models from ShapeNet~\cite{Shapenet} which are typically composed
of semantically labeled parts. For each shape, a hierarchy is constructed with an existing rule-based method~\cite{wang2011symmetry}.
Such principled training hierarchies help the training converges faster.
The loss is computed at each node, including node classification loss and binary point labeling loss.

Our method produces accurate segmentation, even for highly fine-grained decomposition into arbitrary
number of parts, due to the flexibility of dynamic, RvNN-based architecture.
Moreover, it recovers the part relations (adjacency or symmetry) which further improves the labeling accuracy,
e.g., symmetric parts can be identified and thus correctly labeled (Figure~\ref{fig:teaser}).
Our method achieves state-of-the-art performance, both on the public benchmark
and a new benchmark of fine-grained segmentation proposed in this work.
We also demonstrate its utility in image-to-shape reconstruction
with fine-grained structure recovery.

Our contributions include:
\begin{itemize}
\vspace{-6pt}
  \item A deep learning model for top-down hierarchical, fine-grained segmentation of 3D shapes based on dynamic RvNN-based architecture.
\vspace{-6pt}
  \item A part feature learning scheme which integrates both contextual information and per-part shape geometry.
\vspace{-6pt}
  \item A benchmark for fine-grained, part instance segmentation of 3D shapes.
\vspace{-6pt}
  \item An application of our fine-grained structure recovery for high-quality image-to-shape reconstruction.
\end{itemize}



\section{Related work}

\paragraph{Learning 3D shape segmentation.}
Semantic segmentation of 3D shapes has gained significant research progress in recent year, benefiting from the advances in machine learning techniques~\cite{Kalogerakis_SG10,sidi2011unsupervised,wang2013projective,Xie_SGP14,zhao2018triangle}. A comprehensive survey on learning-based 3D shape segmentation can be found in~\cite{xu2016data}.
The basic idea of these approaches is to learn a shape primitive (e.g., a triangle, a point or a voxel) classifier, based on the geometric features of the shape primitives.

Recently, several deep learning models have been developed for supervised segmentation of 3D shapes in various representations including volumetric grid~\cite{riegler2017octnet,Wang2017ocnn}, point cloud~\cite{qi2017pnpp,klokov2017escape,huang2018recurrent}, multi-view rendering~\cite{kalogerakis20173d} or surface mesh~\cite{yi2017syncspeccnn,wang20183d}. The main idea is to replace the hand-crafted geometric features employed in the traditional methods with data-driven learned ones.
All these models, however, are trained targeting a \emph{fixed} set of semantic labels. Given a different set of targeting labels, the model has to be re-trained, using a training dataset annotated with the new labels.

\paragraph{Hierarchical segmentation of 3D shapes.}
3D shapes are usually modeled with parts in a hierarchical construction manner. This is evidenced in part by the wide availability of scene graphs in human-created 3D models of objects or scenes, and by the well-known hierarchical modeling paradigm of Constructive Solid Geometry (CSG)~\cite{requicha1977constructive}. This naturally leads to the idea of hierarchical decomposition of 3D shapes.
Hierarchical shape segmentation can be achieved either with a bottom-up grouping approach~\cite{attene2006hierarchical,wang2018learning}, or in a top-down fashion based on a global topological analysis~\cite{reuter2010hierarchical,huang2009shape,zhou2015generalized}.
Given a pre-segmented 3D shape, Wang et al.~\cite{wang2011symmetry} infer a hierarchical organization of the parts based on proximity and symmetry relations. Later, this heuristic method is improved with an unsupervised learning approach~\cite{van2013co}. Yi et al.~\cite{Yi_SG17} propose a supervised learning approach to hierarchical segmentation of 3D shapes. Their model is, again, trained for a fixed set of semantic tags. The tag sets are determined in a pre-processing of part label analysis and organized with a pre-defined canonical hierarchy. Our method, on the other hand, does not require a prescribed canonical hierarchy and learns the decomposition hierarchies in a data-driven manner, thank to the recursively trained node classification.
Our method is, to our knowledge, the first end-to-end trainable model for hierarchical shape segmentation.

\paragraph{Recursive neural networks.}
Recursive neural nets (RvNN) are developed by Socher et al.~\cite{socher2014recursive}, for text and image understanding~\cite{socher2011parsing}, and for 3D shape classification~\cite{socher2012convolutional}.
Recently, Li et al.~\cite{li2017grass} introduce a generative recursive auto-encoder for generating 3D shape structures. Given a collection of pre-segmented 3D shapes, a variational auto-encoder (VAE) model is learned with RvNN-based encoding and decoding of part structures. Following that, RvNN-based VAE is also trained for 3D scene generation~\cite{li2018grains}, structure-aware single-view 3D reconstruction~\cite{niu2018im2struct} and substructure prior learning for part group composition~\cite{zhu_siga18}.
We are not aware of a previous work on using RvNN for hierarchical 3D shape segmentation.


\section{Method}

We first introduce the overall architecture of PartNet, which is a recursive part decomposition network.
Several key designs in the network will then follow.

\subsection{Recursive part decomposition network}
\begin{figure}[t]
  \centering
  \includegraphics[width=\linewidth]{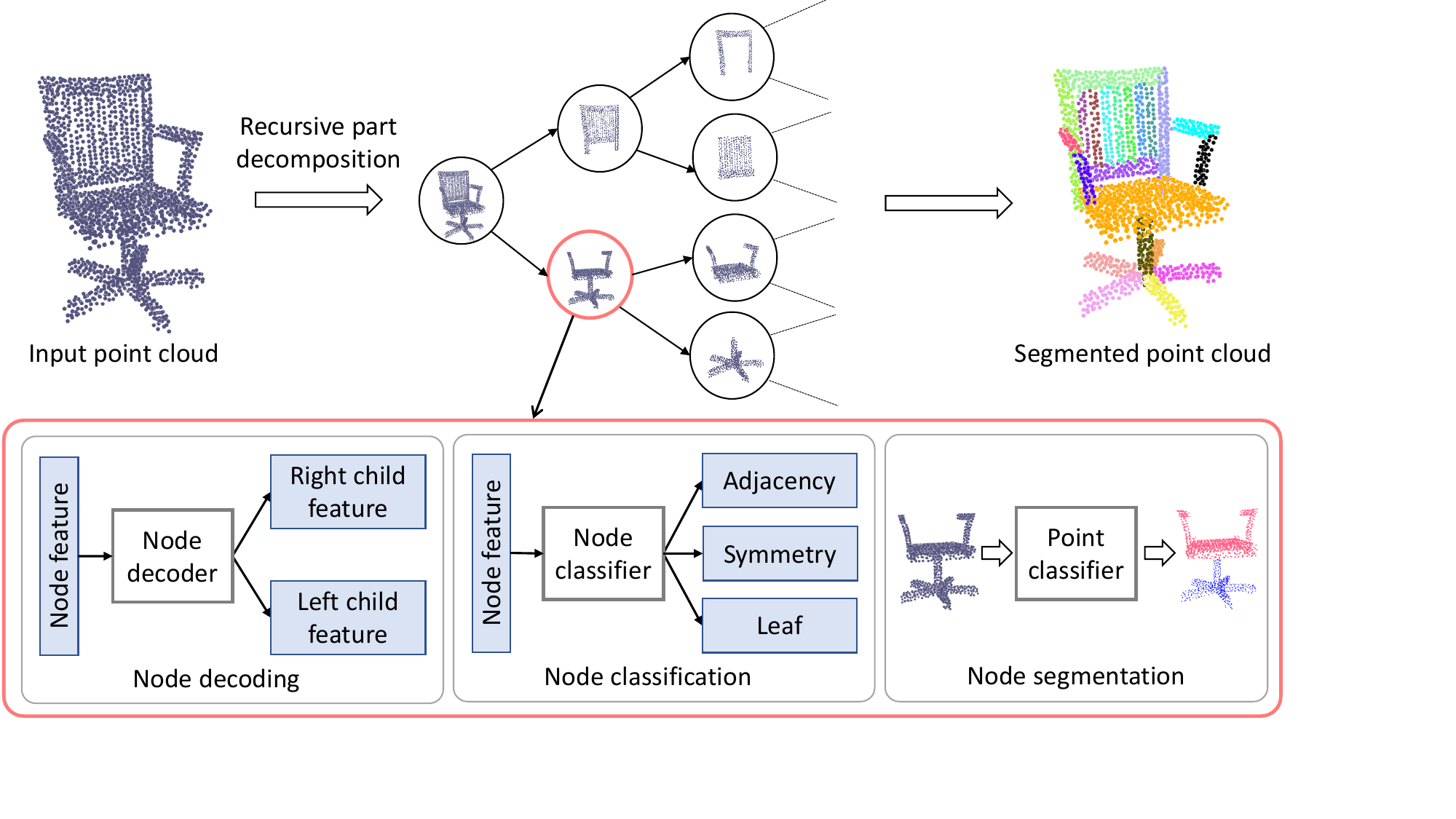}
  \caption{The architecture of PartNet. At each node, there are three modules devised for context propagation, hierarchy construction and point cloud segmentation, respectively. Being a recursive network, these modules are shared by all node in the hierarchy.}
  \label{overview}\vspace{-8pt}
\end{figure}


Figure~\ref{overview} shows the architecture of PartNet. Taking a point cloud of 3D shape as input, PartNet performs a top-down decomposition and and outputs a segmented point cloud at the level of part instances.
At each node, three modules are devised:
\begin{itemize}
  \item \emph{Node decoding module} used to pass the global contextual information from the current node to its children. Such information constraints the segmentation of a node with higher level context.
  \item \emph{Node classification module} devised to construct the topological structure of the decomposition hierarchy. This is achieved by learning to predict the node type which determines how to decompose a node and when to stop the decomposition.
  \item \emph{Node segmentation module} used for performing actual segmentation of the point cloud of the current node. This is achieved by learning a point classification network shared across all nodes.
\end{itemize}
Below we elaborate the discussion on these modules.

\begin{figure}[t]
  \centering
  \includegraphics[width=\linewidth]{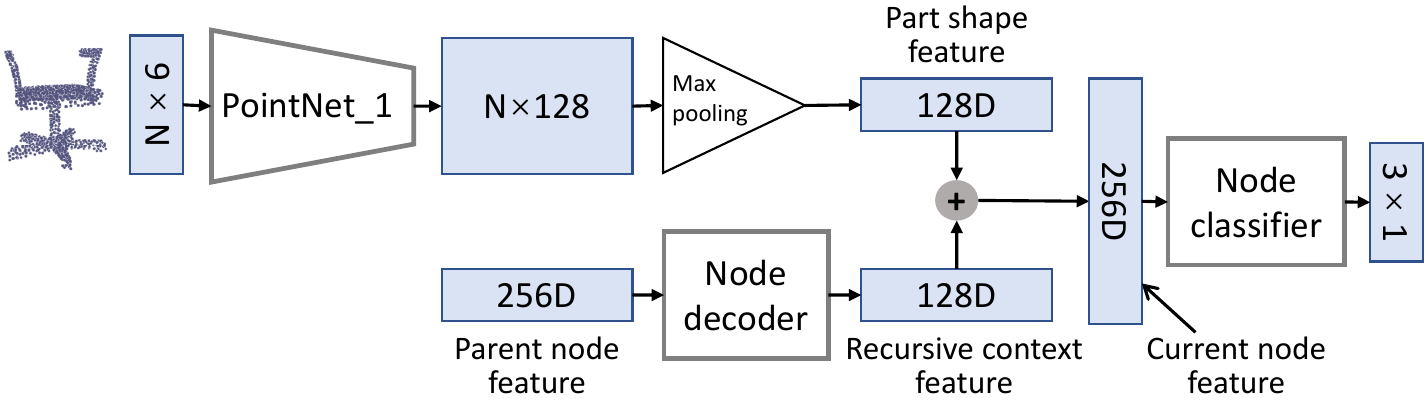}
  \caption{\kx{Network design of the node decoding module and the node classification module. The \emph{recursive contextual feature} and \emph{part shape feature} are concatenated and fed into the node classifier.}}
  \label{fig:node_class}\vspace{-8pt}
\end{figure}

\paragraph{Node decoding module.}
To bootstrap, we first extract a $128$D PointNet~\cite{qi2016pointnet} feature for the full shape point cloud, which is then duplicated and concatenated, forming a $256$D the \emph{root node feature}. This $256$D feature is then decoded into two $128$D features, one for each of its two child nodes, which we refer to as \emph{recursive context feature}.
At each non-root node, we also extract a $128$D PointNet feature for the partial point cloud corresponding to that node, which is called \emph{part shape feature}.
This $128$D part shape feature is then concatenated with the $128$D recursive context feature passed down from the parent node, forming the \emph{current node feature}. Please see Figure~\ref{fig:node_class} for a visual explanation of these features.
%
%
The decoding module is implemented with a two-layer fully connected network with $tanh$ nonlinearity.
This PointNet used in this module is referred to as PointNet\_1, to distinguish with the one to be used in \kx{the node segmentation module (see below).}

\paragraph{Node classification module.}
At a given node, taking its current node feature as input, the node classification module predicts its \emph{node type}
as one of the following three ones: \emph{adjacency}, \emph{symmetry} or \emph{leaf}.
Through determining the how and whether a node is split, this module constructs the topological structure of the hierarchy. This node classification module is implemented with two fully-connected layers with $tanh$ nonlinearity. It can be trained with the ground-truth hierarchical segmentation of a point cloud.

Note that when a node is classified as a symmetry node, we interpret its left child as a symmetry generator (representing a part) and its right child as symmetry parameters, similar to~\cite{li2017grass}. Applying the symmetry parameters on the symmetry generator part obtains the complete point cloud of that symmetry node.
For example, the node corresponding to the spokes in the leg part of a swivel chair (Figure~\ref{fig:teaser}, left) is a rotational symmetry node. Its left child represents the point cloud of one of the spokes and the right child encodes the symmetry axis and symmetry fold.


\begin{figure}[b]
  \centering
  \includegraphics[width=\linewidth]{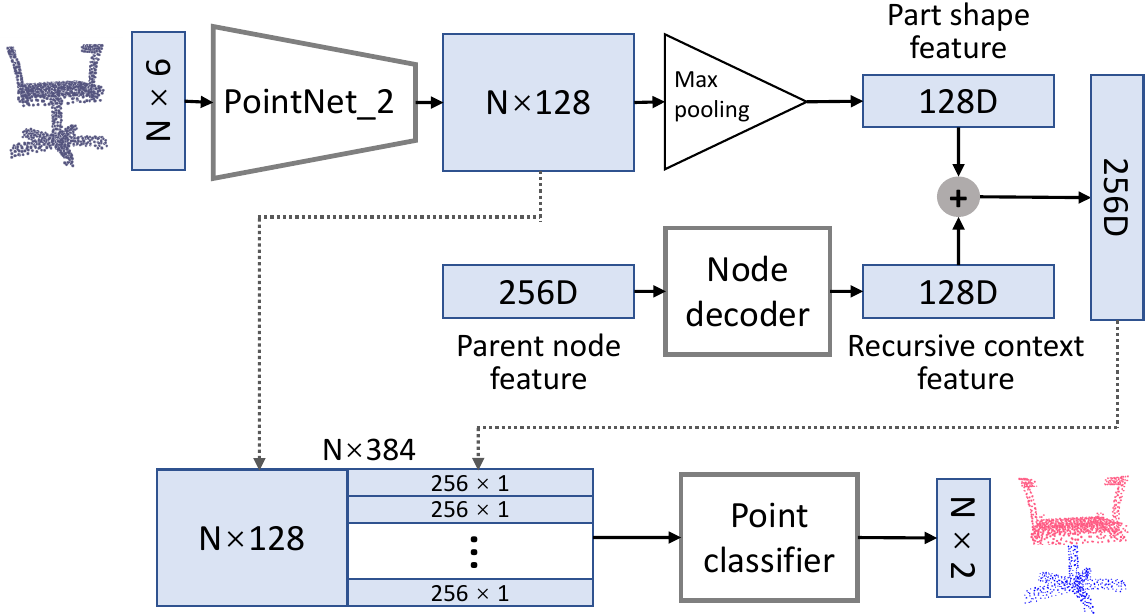}
  \caption{\kx{Network design of the node segmentation module. The concatenation of \emph{recursive contextual feature} and \emph{part shape feature} is enhanced with point-wise PointNet features for the purpose of point label prediction (point cloud segmentation).}}\vspace{-8pt}
  \label{fig:node_seg}
\end{figure}

\paragraph{Node segmentation module.}
This module performs point labeling based on both the current node feature and per-point PointNet features.
Specifically, we use another PointNet (denoted as PointNet\_2) to extract per-point feature, leading to a $N\times 128$ feature matrix, with $N$ being the number of points of the current node.
Then for each row (point feature), we enhance it by concatenating the $256$D current node feature. This results in a $N\times 384$ feature matrix, which is fed into a point classification network to produce point-wise binary labels.
This point classifier is implemented with the last five layers of a PointNet. Note that these layers do not share weight with PointNet\_1 or PointNet\_2.

For a symmetry node, only its left child, i.e., the symmetry generator, needs to be segmented. After the segmentation, the point labels are transferred to all other symmetric counterparts, based on the predicted symmetry parameters.



\subsection{Loss function}
For each training point cloud, the overall loss function for PartNet, ${{L}}_\text{partnet}$, consists of the average node classification loss and average node segmentation loss over all relevant nodes:
\begin{equation}
\label{eq:loss}
{{L}}_\text{partnet} = \frac{1}{|\mathcal{H}|}\sum_{n \in \mathcal{H}}{L_\text{class}(n)} + \frac{1}{|\mathcal{T}|}\sum_{n \in \mathcal{T}}{L_\text{seg}(n)}
\end{equation}
where $L_\text{class}(n)$ and $L_\text{seg}(n)$ are the classification loss and segmentation loss of node $n$, respectively. Both losses are defined as the cross-entropy loss. $\mathcal{H}$ is the set of all nodes in the hierarchy, and $\mathcal{T}$ the set of all non-leaf nodes.


\subsection{Training details}
The PointNet\_1 for node classification (Figure~\ref{fig:node_class}) uses six point convolution layers with $64$, $128$, $128$, $256$, $256$ and $128$ filters, respectively. The PointNet\_2 for node segmentation (Figure~\ref{fig:node_seg}) uses four point convolution layers with $64$, $64$, $128$ and $128$ filters, respectively. The point  cloud segmentation network in Figure~\ref{fig:node_seg} consists of four point convolution layers, with $512$, $256$, $128$ and $128$ filters, respectively, plus the output layer with a $2$ filters for binary label prediction. $20\%$ random feature dropout are used between every two of the last three layers in all these networks. Batch normalization are used between every two layers. We use the Adam optimizer for training, with a batch size of $10$ and the initial learning rate of $0.001$.

The size of input point cloud is $2048$. The training point clouds are obtained by point sampling 3D models. Gaussian noise is added for data enhancement. All PointNets use point normals to improve fine-grained part segmentation performance. Therefore, the dimension of input tensors to PartNet is $2048\times 6$. 


\section{Results and evaluations}

\subsection{Benchmark}

\paragraph{The Fine-grained Segmentation Benchmark (FineSeg).}
With the advances in deep learning based 3D shape segmentation, a benchmark for instance segmentation of fine-grained parts is called for. A nice benchmark for evaluating fine-grained shape segmentation is recently proposed in a concurrent work in~\cite{mo2018partnet}. In this work, we propose FineSeg. The dataset contains about $3000$ 3D shapes over six shape categories: chair ($1000$), table ($500$), airplanes ($600$), sofa ($600$), helicopter ($100$) and bike ($140$). The models are collected from a subset of ShapeNet~\cite{Shapenet} used in the work of Sung et al.~\cite{sung2017}. These models are consistently aligned and uniformly scaled. For those model whose segmentation is not fine-grained enough (e.g., no instance part segmentation), we manually segment the models. We then build a part hierarchy for each shape, using the method proposed in~\cite{wang2011symmetry}.
We point sample each 3D model, thus generating a ground-truth fine-grained segmentation of the corresponding 3D point cloud. The hierarchies can be used for training our recursive part decomposition network. This benchmark can be used to quantitatively evaluate fine-grained segmentation of 3D point clouds, based on \emph{Average Precision (AP)} for part detection (with the IoU against ground-truth greater than a threshold).
The benchmark is publicly available at: \url{www.kevinkaixu.net/projects/partnet.html}.

\begin{figure*}[htbp]
  \centering
  \includegraphics[width=\linewidth]{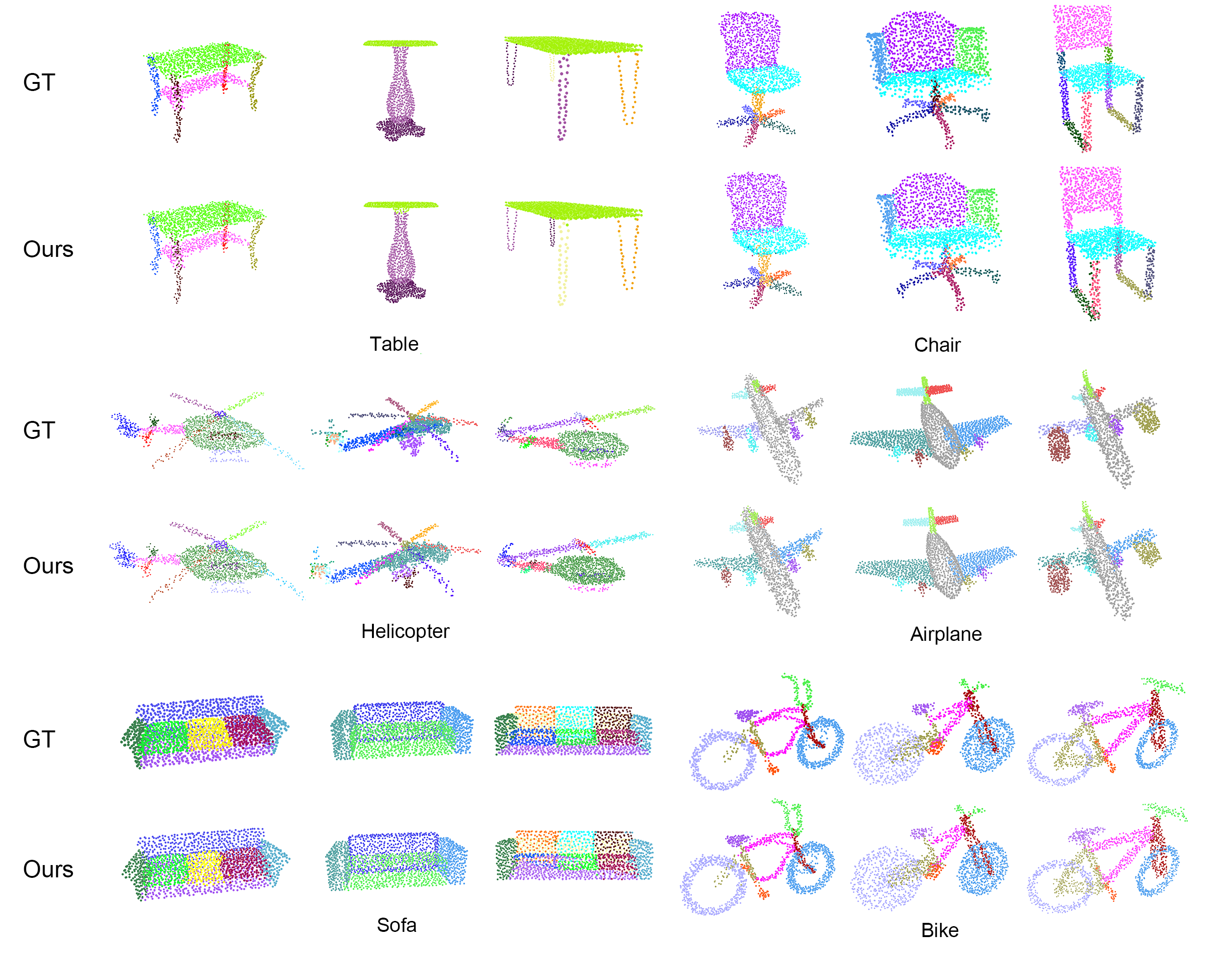}
  \caption{Fine-grained point cloud segmentation by PartNet. For comparison, we show for each shape the fine-grained segmentation result (bottom) and the corresponding ground-truth (top).}
  \label{fine_grained}
\end{figure*}

\subsection{Segmentation results and evaluation}
Our PartNet model is trained with $80\%$ models of FineSeg, leaving the rest $20\%$ for testing.
\kx{The discussion of complexity and timing (for both training and testing) can be found in the supplemental material.}

\begin{figure}[t]
  \centering
  \includegraphics[width=0.8\linewidth]{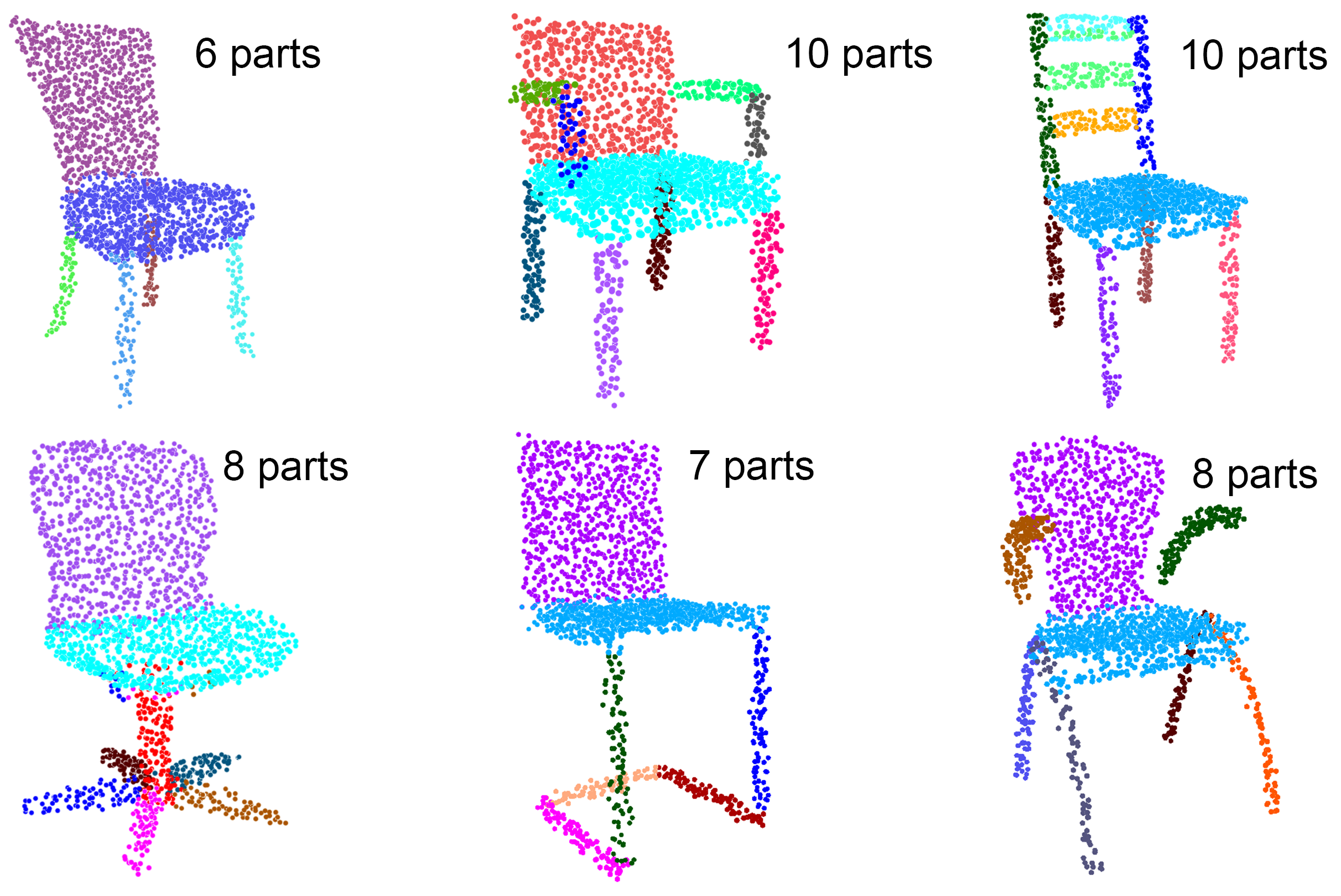}
  \caption{The same PartNet model trained on the Chair set, can be used to segment different chair models into different number of parts, depending on the structure complexity of the input shapes.}
  \label{fig:parts}\vspace{-10pt}
\end{figure}

\paragraph{Visual results on FineSeg.}
We first show in Figure~\ref{fine_grained} some visual examples of fine-grained point cloud segmentation obtained by PartNet. For side-by-side comparison, we also show the ground-truth segmentation for each example. Our method produces precise fine-grained segmentation on the noisy point clouds with complicated part structures. Furthermore, once trained, the same model can be used to segment the test (unseen) point clouds into varying number of parts, demonstrating its flexibility and generality.
Figure~\ref{fig:parts} demonstrates how the same model of PartNet can segment different shapes in a category into for an arbitrary number of targeting parts, depending on structure complexity.
\kx{More results can be found in the supplemental material.}
\kx{In the supplemental material, we also show a visual comparison of hierarchical segmentation
with two traditional (non-learned) baseline methods.}

\paragraph{Quantitative evaluation with ablation study.}\label{ablation study}
In quantitative evaluation on FineSeg, we compare to two baselines which are ablated versions of our method. Specially, we are interested in the effect of the two important node features used in PartNet: recursive context feature (RCF) and part shape feature (PSF).

\begin{table}[!t]\centering\small
\scalebox{0.93}{
\setlength{\tabcolsep}{1.2mm}{
 \begin{tabular}{l|l|c|c|c|c|c|c|c}
\hline
   \multicolumn{2}{c|}{}& mean &aero & bike & chair & heli. & sofa & table \\
\hline
  \multirow{3}*{\makecell[c]{IoU \\ $>0.25$}} & Full & $84.8 $& $95.2$ & $97.0 $& $91.1$ & $83.0 $& $65.4$ &$ 77.2$\\
  &w/o RCF & $79.2$ & $92.8$ & $92.0$ & $87.1$ & $71.1$ &$ 61.6$ & $70.8$\\
  &w/o PSF & $77.6 $& $90.8$ & $95.1$ & $83.6$ &$ 77.8$ & $54.1$ & $64.0$\\
\hline
  \multirow{3}*{\makecell[c]{IoU \\ $>0.5$}} & Full & $72.8$ & $88.0 $& $89.4$ & $80.5$ & $69.4$ &$ 46.7$ &$ 62.6$\\
  &w/o RCF & $66.0 $& $85.3$ &$ 83.4$ & $71.8$ &$ 56.7$ & $42.5 $&$ 56.4$\\
  &w/o PSF & $64.9 $&$ 85.2$ & $88.4 $& $65.6$ & $57.5$ & $36.9$ & $55.6$\\
\hline
\end{tabular}
}}\vspace{5pt}
\caption{Comparing our full model with two baselines (w/o RCF and w/o PSF) on FineSeg. AP($\%$) is measured with IoU threshold being $0.25$ and $0.5$, respectively.}
 \label{ablation_table}\vspace{-6pt}
\end{table}


In the first baseline (w/o RCF), recursive context feature is removed from both node classification (see Figure~\ref{fig:node_class}) and node segmentation (see Figure~\ref{fig:node_seg}). To compensate the missing of recursive context feature, the $128$D part shape feature is duplicated into a $256$D feature. The ablated network is re-trained using the training set of FineSeg.
In the second baseline (w/o PSF), PSF is removed only from the node classification module.

\begin{figure}[t]
  \centering
  \includegraphics[width=\linewidth]{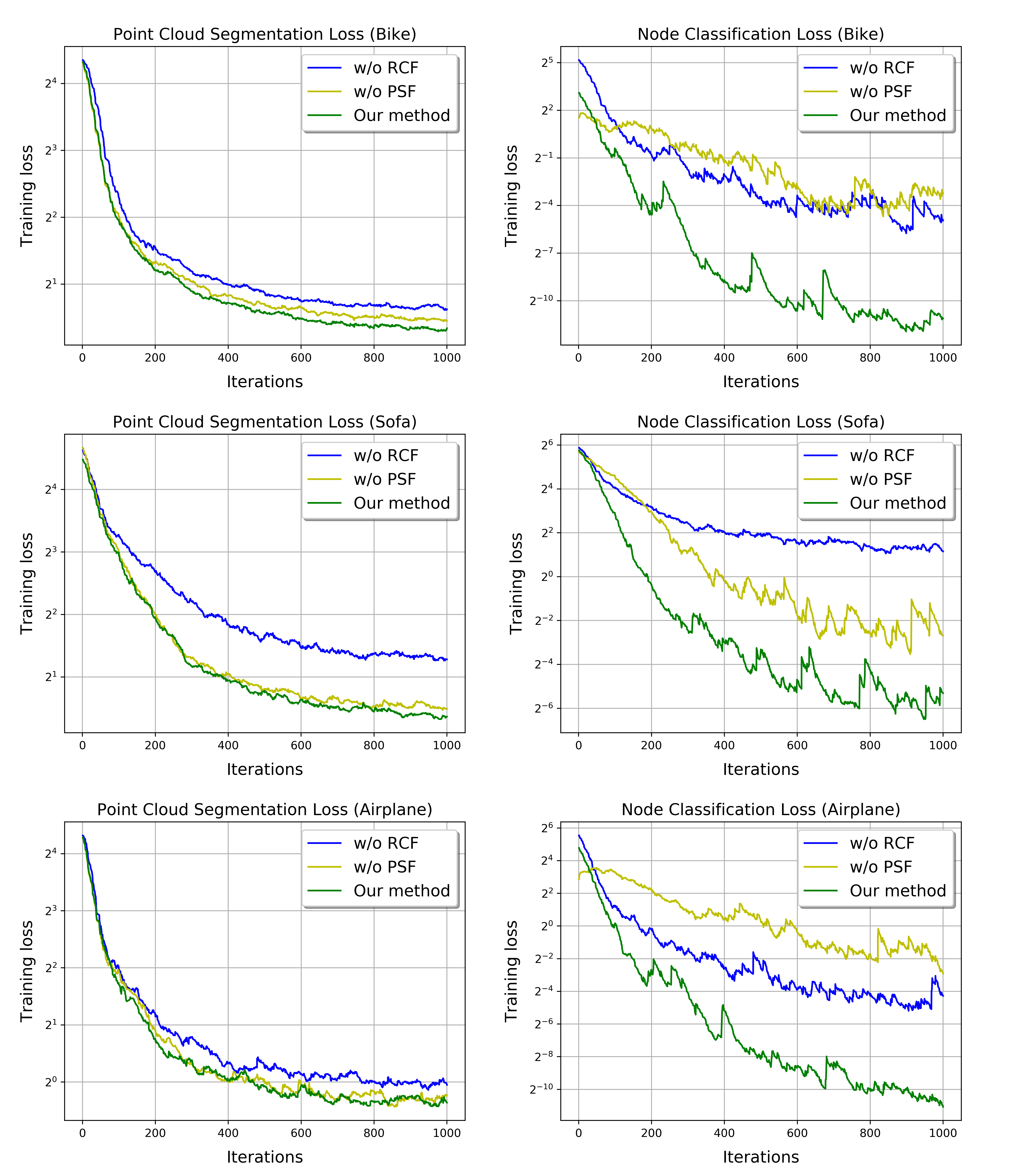}
  \caption{Training loss over iterations in the ablation study of the two key node features (RCF and PSF), on three shape categories (Bike, Sofa and Airplane). For each category (row), we plot both node segmentation loss (left) and node classification loss (right).}
  \label{abalation_figure}
\end{figure}

Table~\ref{ablation_table} reports AP on all six categories of the testing set, with the IoU thresholds being $0.25$ and $0.5$, respectively. The consistent superiority of our full model demonstrates the importance of the two features.
Figure~\ref{abalation_figure} plots the training loss over iterations for the three methods, on three shape categories (Bike, Sofa and Airplane). The results show that both features are critical for fast training of the node classification module and the node segmentation module. This evidences the importance of both global context information and local shape geometry on learning hierarchical segmentation.
\kx{More results are in the supplemental material.}

\paragraph{ShapeNet challenge for fine-grained segmentation.}
In addition, we conduct a moderate-scale stress test using ShapeNet~\cite{Shapenet} challenge for fine-grained segmentation. We randomly select a collection of shapes from ShapNet and use PartNet to segment them.
Since we don't have ground-truth fine-grained segmentation for ShapeNet models, we resort to a subjective study to evaluate our segmentation results. We ask the participants to rate the quality of fine-grained segmentation in the range from $1$ to $5$. \kx{The user study shows that our method attains $>4.0$ average ratings for all the categories tested, much higher than the results of the ``w/o RCF'' baseline.} \kx{The details and results of this study are provided in the supplemental material.} In Figure~\ref{shapenet}, we show a few visual examples, from which one can see that our method produces fine-grained segmentation these unseen shapes with complicated structures. Moreover, our method obtains the adjacency and symmetry relations of the decomposed parts, which can be used for many downstream structure-aware applications~\cite{mitra2013structure}.


\begin{figure}[t]
  \centering
  \includegraphics[width=\linewidth]{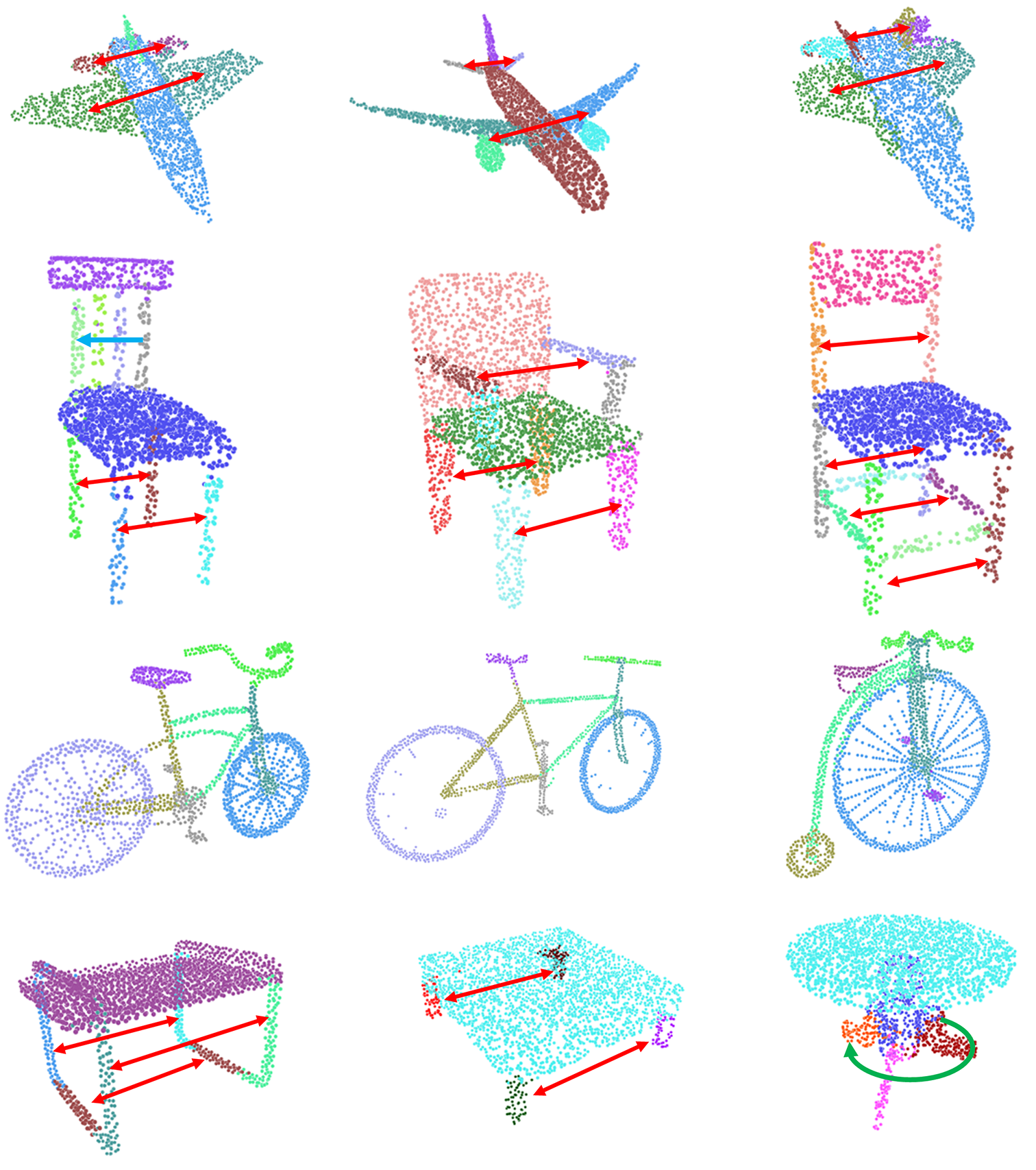}
  \caption{A few results from the ShapeNet fine-grained segmentation challenge. Besides segmentation, PartNet can also recover the relations (adjacency and symmetry) between the segmented parts. We visualize the recovered symmetry relations with colored arrows (Reflective: Red; Translational: Blue; Rotational: Green).}
  \label{shapenet}\vspace{-8pt}
\end{figure}

\subsection{Comparison of semantic segmentation}
\label{subsec:semantic}
Although PartNet is designed for fine-grained segmentation, the recursive decomposition should work even better for semantic segmentation since the latter is usually a much coarser-level segmentation. We evaluate PartNet for semantic segmentation of 3D point clouds on the ShapeNet part dataset~\cite{Yi16}, through comparing with seven state-of-the-art methods on this task. Similar to PointNet~\cite{qi2016pointnet}, we re-sample the point cloud for each shape into $2048$ points. We use the same training/testing split setting as those state-of-the-arts, and compute part-wise average IoU as metric.

Note that PartNet does not produce semantic labels for points, so it cannot perform labeled segmentation. To enable the comparison, we add an extra module to PartNet to predict a semantic label for each part it decomposes. The part label prediction module takes the node feature of the leaf nodes as input and outputs a semantic label for all points included in that leaf node. This module is implemented with three fully-connected layers and is trained with cross-entropy loss.

\begin{table*}[!t]\centering\small
 \scalebox{0.98}{
\setlength{\tabcolsep}{1.2mm}{
 \begin{tabular}{l|c|c|c|c|c|c|c|c|c|c|c|c|c|c|c|c|c}
\hline
  Method & mean &aero & bag & cap & car & chair & eph. & guitar & knife & lamp & laptop & motor & mug & pistol & rocket & skate. & table\\
\hline
  PointNet~\cite{qi2016pointnet}&$83.7$&$83.4$&$78.7$&$82.5$&$74.9$&$89.6$&$73.0$&$91.5$&$85.9$&$80.8$&$95.3$&$65.2$&$93.0$&$81.2$&$57.9$&$72.8$&$80.6$\\
  PointNet++~\cite{qi2017pnpp}&$85.1$&$82.4$&$79.0$&$87.7$&$77.3$&$90.8$&$71.8$&$91.0$&$85.9$&$83.7$&$95.3$&$71.6$&$94.1$&$81.3$&$58.7$&$76.4$&$82.6$\\
  O-CNN~\cite{Wang2017ocnn}&$85.9$&$85.5$&\textbf{87.1}&$84.7$&$77.0$&$91.1$&\textbf{85.1}&$91.9$&$87.4$&$83.3$&$95.4$&$56.9$&$96.2$&$81.6$&$53.5$&$74.1$&$84.4$\\
  SSCN~\cite{graham20183d}&$86.0$&$84.1$&$83.0$&$84.0$&\textbf{80.8}&$91.4$&$78.2$&$91.6$&\textbf{89.1}&$85.0$&$95.8$&$73.7$&$95.2$&$84.0$&$58.5$&$76.0$&$82.7$\\
  PCNN~\cite{PCNN18}&$85.1$&$82.4$&$80.1$&$85.5$&$79.5$&$90.8$&$73.2$&$91.3$&$86.0$&$85.0$&$95.7$&$73.2$&$94.8$&$83.3$&$51.0$&$75.0$&$81.8$\\
  SPLATNet~\cite{su18splatnet}&$85.4$&$83.2$&$84.3$&$89.1$&$80.3$&$90.7$&$75.5$&$92.1$&$87.1$&$83.9$&$96.3$&$75.6$&$95.8$&$83.8$&$64.0$&$75.5$&$81.8$\\
  PointCNN~\cite{NIPS2018_7362}&$86.1$&$84.1$&$86.4$&$86.0$&\textbf{80.8}&$90.6$&$79.7$&\textbf{92.3}&$88.4$&\textbf{85.3}&$96.1$&\textbf{77.2}&$95.3$&\textbf{84.2}&$64.2$&\textbf{80.0}&$83.0$\\
\hline
 Ours&\textbf{87.4}&\textbf{87.8}&$86.7$&\textbf{89.7}&$80.5$&\textbf{91.9}&$75.7$&$91.8$&$85.9$&$83.6$&\textbf{97.0}&$74.6$&\textbf{97.3}&$83.6$&\textbf{64.6}&$78.4$&\textbf{85.8}\\
\hline
\end{tabular}}}\vspace{5pt}
\caption{Comparison of semantic segmentation on the ShapeNet part dataset~\cite{Yi16}. Metric is part-wise IoU ($\%$).}
\label{shapenet_table}\vspace{-8pt}
\end{table*}

The results are reported in Table~\ref{shapenet_table}. PartNet, augmented with a part label prediction module, achieves better performance in most of the categories, and the highest mean accuracy over all categories. Furthermore, our method works especially well for those categories with complex structures such as chair, table, and aeroplane, etc. We believe that the divide-and-conquer nature of recursive decomposition does help reduce the difficulty of segmentation learning. Another key benefit of recursive decomposition is that the segmentation of higher levels provides contextual cues constraining that of the lower levels.
\kx{Similar results can also be observed in testing our trained model on
the Princeton Segmentation Benchmark~\cite{Chen:2009:ABF} (see supplemental material).}

\kx{
For semantic segmentation, PartNet can be trained with a consistent hierarchy for all shapes in a category.
The training is can be done with \emph{any} hierarchy that is consistent across all training shapes.
Therefore, we do \emph{not} need an extra process (such as the one~\cite{wang2011symmetry} used in fine-grained segmentation) for hierarchy construction. Taking \emph{any random hierarchy} of one training shape as a ``template'', we unify the hierarchies of all the other shapes based on the semantic part labels.
Therefore, PartNet does \emph{not} require an extra supervision of part hierarchy for training for semantic segmentation. Consequently, the comparison reported in Table~\ref{shapenet_table} of the main paper is a fair one.
}

\subsection{Comparison of instance segmentation}
SGPN~\cite{Wang2017SGPN} is the first deep learning model that learns instance segmentation on 3D point clouds. It can segment object instances and predict a class label for each instance, which is very similar to our method (augmented the label prediction module), except that SGPN cannot obtain part relations as our method does. We make a comparison to SGPN on our FineSeg dataset, using again AP with IoU thresholds of $0.25$ and $0.5$.

\begin{figure}[t]
  \centering
  \includegraphics[width=\linewidth]{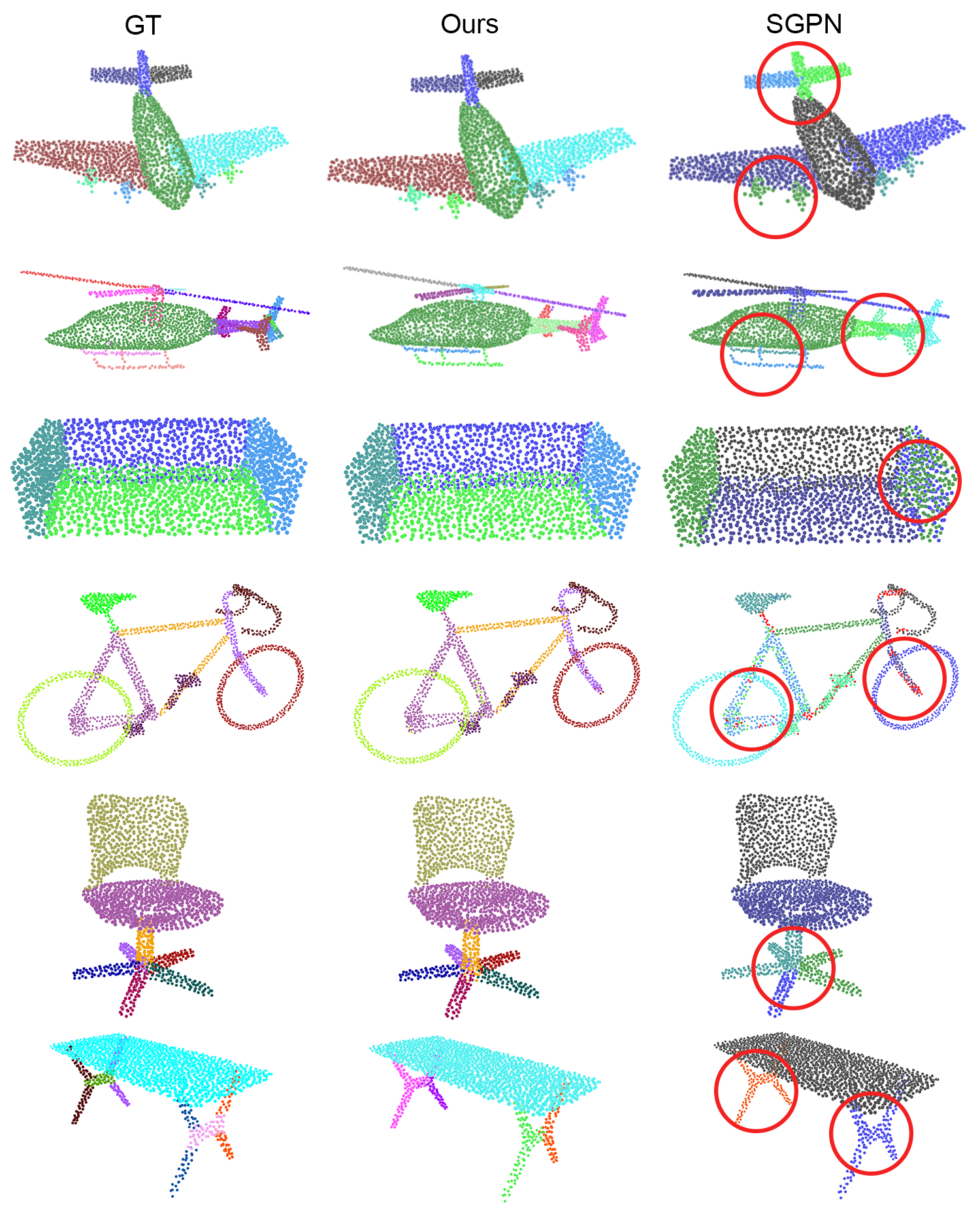}
  \caption{Visual comparison of fine-grained part instance segmentation with SGPN~\cite{Wang2017SGPN}. Left: Ground-truths. Middle: Segmentation results by PartNet. Right: Results by SGPN. Incorrect segmentations (w.r.t. ground-truth) are marked with red circles.}
  \label{sgpn_figure}\vspace{-5pt}
\end{figure}

\begin{table}[!t]\centering\small
 \scalebox{0.98}{
\setlength{\tabcolsep}{1.0mm}{
 \begin{tabular}{c|l|c|c|c|c|c|c|c}
\hline
   \multicolumn{2}{c|}{}& mean &aero & bike & chair & heli. & sofa & table \\
\hline
  \multirow{2}*{\makecell[c]{IoU \\ $>0.25$}} & SGPN~\cite{li2017grass} & $62.2$ & $67.8$ & $75.8$ & $66.2$ & $59.4$ & $50.4$ & $53.6$\\
  & Ours & $84.8$ & $95.2$ & $97.0$ & $91.1$ & $83.0$ & $65.4$ & $77.2$\\
\hline
  \multirow{2}*{\makecell[c]{IoU \\ $>0.5$}} & SGPN~\cite{li2017grass} & $47.0$ & $56.7$ & $63.7$ & $54.6$ & $38.9$ & $29.5$ & $38.4$\\
  & Ours & $72.8$ & $88.0$ & $89.4$ & $80.5$ & $69.4$ & $46.7$ & $62.6$\\
\hline
\end{tabular}}}\vspace{5pt}
\caption{Comparison with SGPN~\cite{Wang2017SGPN} on fine-grained instance segmentation over the FineSeg dataset. The metric is AP ($\%$) with IoU threshold being $0.25$ and $0.5$, respectively.}
\label{sgpn_table}\vspace{-8pt}
\end{table}

Figure~\ref{sgpn_figure} shows a few visual comparisons, where incorrectly segmented regions are marked out. Table~\ref{sgpn_table} shows the quantitative comparison on our datasets. We attribute the consistent improvement over SGPN to two factors. First, the instance group learning of SGPN is based on point clustering. Such a one-shot point grouping over the entire shape is hard to learn. Our method, on the other hand, performs top-down recursive decomposition which breaks the full shape segmentation into a cascade of partial shape segmentations. Second, the point features used by SGPN are solely point convolutional features~\cite{qi2016pointnet} while our features accounts for both local part shape and global context.

\section{Applications}
We demonstrate an application of the fine-grained segmentation of PartNet in refining 3D point clouds reconstructed from single view images. The basic idea is a \emph{segment-and-refine} process. Given a 3D point cloud reconstructed in a holistic fashion (using, e.g., the method of Fan et al.~\cite{Fan2017A}), we first perform a recursive decomposition of the point cloud, resulting in a hierarchical organization of part point clouds.
We then train a network to refine the part point cloud at each leaf node, yielding a high-quality point cloud for that part. These refined part point clouds together constitute a refined point cloud of the full shape.


The part refiner network used at each leaf node is composed of two channels of PointNet, to encode the point clouds of the part and the full shape, respectively. The resulting two features are concatenated and fed into a four layer fully-connected networks to generate a refined part point cloud. To train this refiner network, we use reconstruction loss computed as the Chamfer distance and the earth mover's distance between point clouds~\cite{Fan2017A}. To gain more training signals, we opt to train the refiner with a hierarchical reconstruction loss, through a bottom-up composition of the refined part point clouds, following the hierarchy obtained by PartNet segmentation. This way, we can compute a reconstruction loss at each node of the hierarchy, with the corresponding point cloud composed from the part point clouds within its subtree. \kx{Please refer to the supplemental material for more details on the network architecture.}

\begin{figure}[t]
  \centering
  \includegraphics[width=\linewidth]{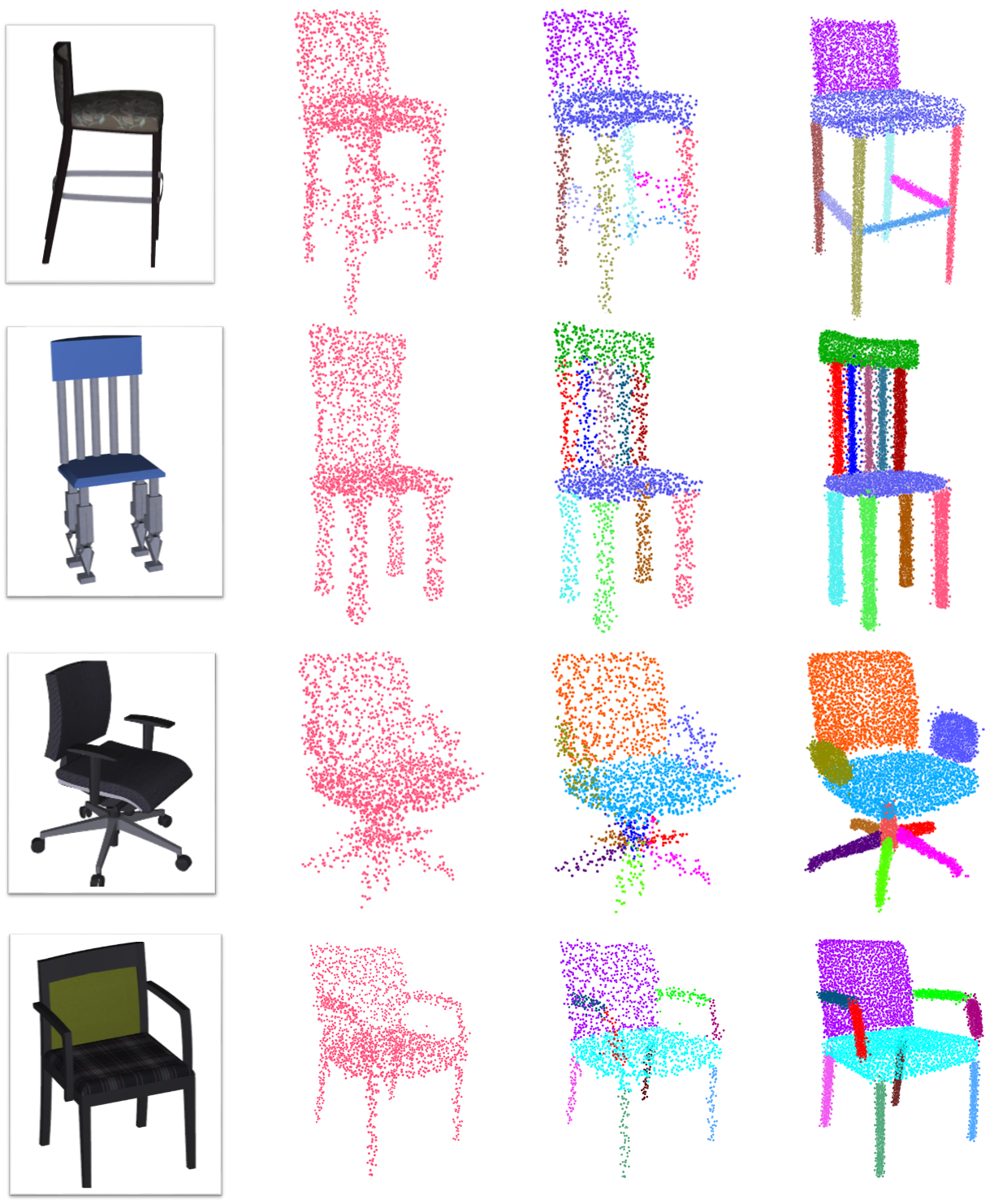}
  \caption{A few examples on refining point clouds reconstructed from single view images, guided by the fine-grained segmentation of PartNet. In each row, we show from left to right the input image, result of holistic reconstruction, fine-grained segmentation of the reconstruction, and the final refinement result by our method. }
  \label{app}\vspace{-8pt}
\end{figure}

Figure~\ref{app} shows a few examples of point cloud refinement, guided by the fine-grained, hierarchical segmentation of PartNet. Although the refinement may sometimes lose part fidelity w.r.t. the input images, it does produce highly detailed point clouds and with plausible part structure, thanks to the fine-grained part decomposition of PartNet. \kx{See more examples in the supplemental material.}



\section{Conclusion}

We have presented a top-down recursive decomposition network for fine-grained segmentation of 3D point clouds. With the hierarchical decomposition scheme, our model obtains fine-grained and accurate segmentation even for highly complex shapes. Different from most existing deep-learning based segmentation models, our method segments a shape into an arbitrary number of parts, depending on its structural complexity, instead of producing a labeled segmentation with a fixed label set. Even for semantic segmentation, our model also achieves superior performance, benefiting from our divide-and-conquer segmentation learning.

\paragraph{Limitations and future work.}
Our current method has a few limitations.
\emph{First}, although PartNet segments a shape in a hierarchical fashion, the resulting segment hierarchy is not necessarily as meaningful as the those learned purposively for shapes~\cite{van2013co,Yi_SG17} or scenes~\cite{liu2014creating}.
\emph{Second}, although our model can be used to segment different shapes into different number of parts, instead of targeting a fixed part label set. It still needs to be trained for each shape category separately. Learning a more general model for recursive decomposition of shapes from multiple classes would be a very interesting future direction to look into.
\emph{Third}, PartNet is trained with reasonable ground-truth hierarchies built with the method in~\cite{wang2011symmetry}. Training with totally random hierarchy would lead to performance degrading. \kx{We show this effect in the supplemental material.} Therefore, another future direction is to learn hierarchical segmentation in a unsupervised manner, without the need of building ground-truth hierarchies.

\section*{Acknowledgement}
We thank the anonymous reviewers for their valuable comments. We are also grateful to Hao (Richard) Zhang for the fruitful discussions, and to Yuan Gan and Pengyu Wang for the tremendous help on data preparation. This work was supported in part by NSFC (61572507, 61532003, 61622212) and Natural Science Foundation of Hunan Province for Distinguished Young Scientists (2017JJ1002).

{\small
\bibliographystyle{ieee}
\bibliography{partnet}
}

\end{document}